\documentclass{article}

\usepackage{PRIMEarxiv}
\usepackage{tabularx}
\usepackage[utf8]{inputenc} 
\usepackage[T1]{fontenc}    
\usepackage{hyperref}       
\usepackage{url}            
\usepackage{booktabs}       
\usepackage{amsfonts}       
\usepackage{nicefrac}       
\usepackage{microtype}      
\usepackage{lipsum}
\usepackage{fancyhdr}       
\usepackage{graphicx} 
\usepackage{amsmath} 
\graphicspath{{media/}}     

\title{When to Switch, Not Just What: Transition Quality Prediction in Clash Royale\\
}

\author{
Heeyun Heo\\
School of Cybersecurity\\
Korea University\\
Seoul, Republic of Korea\\
\texttt{heeyun0724@korea.ac.kr}
\and
Huy Kang Kim\\
School of Cybersecurity\\
Korea University\\
Seoul, Republic of Korea\\
\texttt{cenda@korea.ac.kr}
}

\begin{document}
\maketitle

\begin{abstract}
In competitive games, players frequently switch strategies after losing streaks, yet our analysis of 926,334 match records from 34,619 \emph{Clash Royale} players reveals a counterintuitive pattern: switching frequency is inversely associated with the win rate, with effects that vary substantially across players and situational contexts. We attribute this to a limitation common in many prior recommendation systems, which evaluate strategies by expected quality while overlooking the behavioral cost of switching and individual differences in switching propensity. We refer to this implicit premise as the \emph{Zero Switching Cost Assumption}. To address this, we reformulate strategy recommendation as a \emph{transition-level decision problem} and instantiate it as TQP (Transition Quality Predictor), a three-stage pipeline structured as \textsc{Who}~$\rightarrow$~\textsc{When}~$\rightarrow$~\textsc{What}. PersonaGate suppresses recommendations for players whose strategic consistency is empirically associated with superior outcomes. TimingGate identifies moments when switching is likely to yield a net benefit over staying, using a subtype- and state-matched baseline to control for natural win-rate recovery. ScoreFusion ranks candidate strategies by combining an adoptability signal with predicted transition quality ($\Delta WR$). We further introduce SwitchGap, an evaluation metric that measures a policy's discriminative quality without treating observed player choices as optimal ground truth. This property is particularly important because the most frequent switchers record the lowest win rates. The full pipeline achieves a SwitchGap of $+10.4$\%p at a recommendation rate of $5.4\%$, and loss-triggered switchers, despite being the lowest-performing group, benefit the most from subtype-conditioned guidance.
\end{abstract}


\section{Introduction}
\label{sec:intro}
In competitive online games, players often switch strategies after a sequence of losses. Prior work has shown that consecutive defeats often trigger coping behaviors such as strategy switching, character switching, and disengagement~\cite{wu2021, deng2024}. In meta-volatile environments like \emph{Clash Royale}, where frequent balance updates reshape the competitive landscape, exploring new strategies can appear to be a rational response.

Yet our behavioral log analysis reveals a counterintuitive pattern: the most frequent switchers tend to achieve \emph{lower} average win rates than those who maintain strategic consistency. Moreover, the magnitude of this effect varies markedly across players' behavioral profiles, suggesting that \emph{who} receives a transition recommendation and \emph{when} it is made are likely key factors.

We interpret this heterogeneity through the lens of \emph{Adaptation Cost}, a construct we use to describe two primary burdens we associate with strategy switching. Motivated by prior evidence that game-specific expertise materially affects match outcomes~\cite{do2021}, we define \emph{Mastery Loss} as the transient performance degradation that may arise before a player becomes proficient with a newly adopted strategy. In this study, this possibility is addressed implicitly through the net transition effect formulation, which controls for expected win-rate recovery under matched conditions.

We define \emph{Tilt-like Reactivity} as a set of behaviorally observable reactive responses in post-loss contexts, including abrupt switching and other erratic or non-deliberate adjustments. While this concept is broadly consistent with prior descriptions of tilt, our formulation infers it from behavioral signals rather than direct psychological measurement~\cite{wu2021, kou2020, kou2018, deng2024}. Consistent with this interpretation, the win-rate gaps observed across behavioral groups suggest that the same switching behavior can lead to markedly different outcomes depending on a player's subtype.

Despite the significance of this pattern, prior research leaves two important gaps. First, much prior work has focused on diagnosis---that is, retrospective behavioral classification---rather than prescription: actionable guidance at the moment a decision must be made. Second, many recommendation systems do not explicitly model the intrinsic costs of switching. When these costs are overlooked, systems may recommend transitions that appear theoretically attractive but are poorly timed for the current player. In adjacent recommender settings, unsatisfactory recommendations have been shown to increase churn risk under uncertainty~\cite{keinan2025}. We refer to this implicit premise as the \emph{Zero Switching Cost Assumption}: the assumption that any player can transition instantaneously and without penalty, regardless of mastery degradation, loss-induced reactivity, or individual switching propensity.

The fundamental research question is therefore not merely \emph{whether} to switch, but \emph{for whom} and \emph{at what moment} a transition yields a net improvement. We operationalize this question by comparing the outcome of a transition against the win-rate change observed among matched players who maintained their current strategy, thereby controlling for the recovery that might occur even without a strategic change.

To address these limitations, we reformulate strategy recommendation as a \emph{transition-level decision problem} under behavioral constraints, and instantiate this formulation as TQP (\textbf{T}ransition \textbf{Q}uality \textbf{P}redictor), a three-stage pipeline structured as \textsc{Who}~$\rightarrow$~\textsc{When}~$\rightarrow$~\textsc{What}. PersonaGate suppresses recommendations for players whose behavioral consistency is empirically associated with superior outcomes. TimingGate identifies moments when switching is likely to yield a net benefit over staying. ScoreFusion then ranks candidate strategies by combining an adoptability signal with predicted transition quality ($\Delta WR$).

\section{Related Work}
\subsection{Player Behavioral Heterogeneity and Strategy Adaptation}

Competitive game players exhibit systematic behavioral heterogeneity: individuals differ not only in skill level but also in how they respond to in-game events and adapt their strategies over time. Ingram \textit{et al.}~\cite{ingram2022} demonstrated that trajectory-based behavioral signals can identify distinct play styles through an unsupervised LSTM-autoencoder clustering approach, showing that such heterogeneity is both measurable and persistent across gameplay sessions. Building on this line of work, Ingram \textit{et al.}~\cite{ingram2025} paired each identified style with a style-specific expert policy to generate play-style-conditioned advice, improving simulated player performance when the advice is followed. Subsequent work further reported statistically significant performance gaps across behaviorally defined subgroups~\cite{elbert2024, teng2023}. More broadly, performance-based clustering studies in competitive games have shown that behavioral groups diverge markedly in their performance trajectories over time~\cite{qiu2024, sembina2025}. Taken together, these findings suggest that behavioral subgroup membership is not merely descriptive of play style, but is also associated with meaningful differences in player outcomes.

However, existing research has predominantly examined player performance \emph{within} relatively stable strategic settings, while giving less attention to the strategy transition decision itself as a primary unit of analysis. Even play-style-conditioned advice frameworks typically determine advice by comparing current behavior with an expert model, without estimating the behavioral cost of acting on it or whether the change is likely to outperform staying. As a result, the possibility that the effectiveness of a transition depends on the player's behavioral profile, that is, the \emph{who} behind the switch, has remained underexplored in recommendation design. Under the \emph{Adaptation Cost} lens introduced in Section~\ref{sec:intro}, the same transition recommendation may therefore yield substantially different outcomes across behavioral subtypes.

\subsection{Implicit Costless-Switching Assumptions in Recommendation Systems}

Strategy recommendation research has evolved from optimizing team synergies to user profiling based on card preferences and historical win rates~\cite{chen2018, lee2022}, with recent frameworks attempting to capture transition patterns through meta-learning or frequency-based profiling~\cite{wang2021, zang2022}. Beyond game-specific settings, behavior-aware recommender systems such as ATRank~\cite{zhou2018atrank} modeled users through their historical action sequences via self-attention. However, these approaches primarily rank candidate items or actions rather than estimating the transition-level outcome of moving from one strategy to another.

A related form of the \emph{when-to-switch} problem has been studied in reinforcement learning, where Comanici and Precup~\cite{comanici2010} learned termination conditions over pre-defined policies to switch an agent between them. Their formulation, however, optimizes an agent's long-term return rather than modeling human switching behavior. In contrast, our work treats the strategy transition itself as the unit of recommendation and predicts whether a specific switch yields a positive net effect on subsequent win rate, relative to staying.

Despite these advances, these systems implicitly adopt the Zero Switching Cost Assumption (Section~\ref{sec:intro}) and focus on identifying a theoretically superior strategic destination rather than modeling the actual win rate change ($\Delta WR$) experienced during a transition. Consequently, applying a uniform recommendation logic to all players risks inducing unnecessary transitions and adverse outcomes.

\section{Dataset and Behavioral Analysis}
\label{sec:analysis}
\subsection{Data Collection}
\label{sec:data}
This study uses match-level logs collected through the public API of the mobile card battle game \textit{Clash Royale}~\cite{clashroyale_api}. Our analysis focuses on competitive modes involving real-time one-on-one combat and ladder progression, specifically Player-versus-Player (PvP) and Path of Legend. To ensure statistical robustness, we applied several exclusion criteria. We excluded players with fewer than 20 total matches. Furthermore, we removed players with fewer than five recorded observations following a loss or fewer than five following a win, in order to maintain the reliability of the behavioral analysis. After applying these filters, the final analytical sample consisted of 34,619 players.

\subsection{Strategy State Definition}

\begin{table}[t]
\centering
\small
\caption{Strategy State Definitions}
\label{tab:states}
\begin{tabular*}{0.65\textwidth}{@{\extracolsep{\fill}}ll@{}}
\toprule
\textbf{Group} & \textbf{Name} \\
\midrule
Cycle      & Hyper Cycle (7) \\
           & Classic Cycle (4) \\
           & Off-Meta Cycle (11) \\
\addlinespace
Control    & Unit-Heavy Control (1) \\
           & Standard Control (2) \\
           & Bridge Spam (3) \\
\addlinespace
Beatdown   & Defensive Heavy Variant (10) \\
           & Defensive Heavy (6) \\
           & Siege / Heavy Control (9) \\
           & Classic Beatdown (5) \\
           & All-In Beatdown (0) \\
\addlinespace
Specialist & Three Musketeers (8) \\
           & Spell Cycle / Troll (12) \\
\bottomrule
\end{tabular*}
\end{table}

To quantify the strategic space reflected in players' deck compositions, we performed archetype clustering using seven structural indicators, including average elixir cost, cost standard deviation, and four functional card-type ratios. After normalization with a QuantileTransformer, we applied domain-specific weights and derived 13 strategic states ($k=13$, $Silhouette=0.54$). High clustering stability was confirmed through repeated experiments, yielding an ARI of $0.88$ and an NMI of $0.95$. The resulting 13 clusters correspond to major in-game archetype categories such as Cycle, Control, and Beatdown. These clusters serve as the state units that define the origins and destinations of strategy transitions within the TQP pipeline (Table~\ref{tab:states}).

\subsection{Player Behavioral Subtypes}
\label{sec:subtypes}

To capture players' strategic volatility, we performed KMeans clustering ($k=3$) along three primary axes: Consistency, Dynamics, and Reactivity. The key behavioral features underlying these axes are defined as follows. \texttt{overall\_switch\_rate} denotes the proportion of matches in which a player changed their deck compared with the previous match. \texttt{loss\_reactivity} is defined as the difference between the post-loss deck change rate and the post-win deck change rate, quantifying the asymmetric sensitivity of switching behavior to defeat. \texttt{avg\_change\_magnitude} measures the average structural extent of deck changes, computed as the mean Jaccard distance between consecutive decks across all matches in which a change occurred. Notably, \texttt{loss\_reactivity} serves as a key feature for distinguishing between a general preference for strategic diversification and loss-triggered switching behavior consistent with tilt-like reactivity. Validated through silhouette analysis (Silhouette$=0.72$) and iterative relabeling (ARI$=0.997$, NMI$=0.993$), this classification framework provides a stable representation of behavioral heterogeneity among players. This stability provides a robust foundation for subsequent persona-conditioned recommendations within the TQP pipeline. 

\begin{table}[h]
\centering
\caption{Behavioral Clustering Axes and Definitions}
\label{tab:subtype_axes}
\small
\begin{tabularx}{\columnwidth}{lX}
\toprule
\textbf{Analysis Axis} & \textbf{Role and Definition} \\
\midrule
Consistency & Measures strategic consistency via the occupancy rate of the most frequently used deck, along with auxiliary 
diversity indicators. \\
\addlinespace
Dynamics    & Captures transition frequency and the structural 
magnitude of changes. \\
\addlinespace
Reactivity  & Quantifies asymmetric sensitivity to defeat via 
\texttt{loss\_reactivity}, serving as the primary discriminating feature across subtypes. \\
\bottomrule
\end{tabularx}
\end{table}

\subsection{Behavioral Differentiation and Performance}

Clustering results identified three behaviorally distinct player subtypes that were clearly separated along the \texttt{loss\_reactivity} axis. This suggests that responses to failure constitute a major dimension of behavioral differentiation. The central finding of this analysis is a counterintuitive pattern: more frequent switching is associated with lower performance, contrary to common intuition.

\begin{figure}[h]
    \centering
    \includegraphics[width=0.75\textwidth]{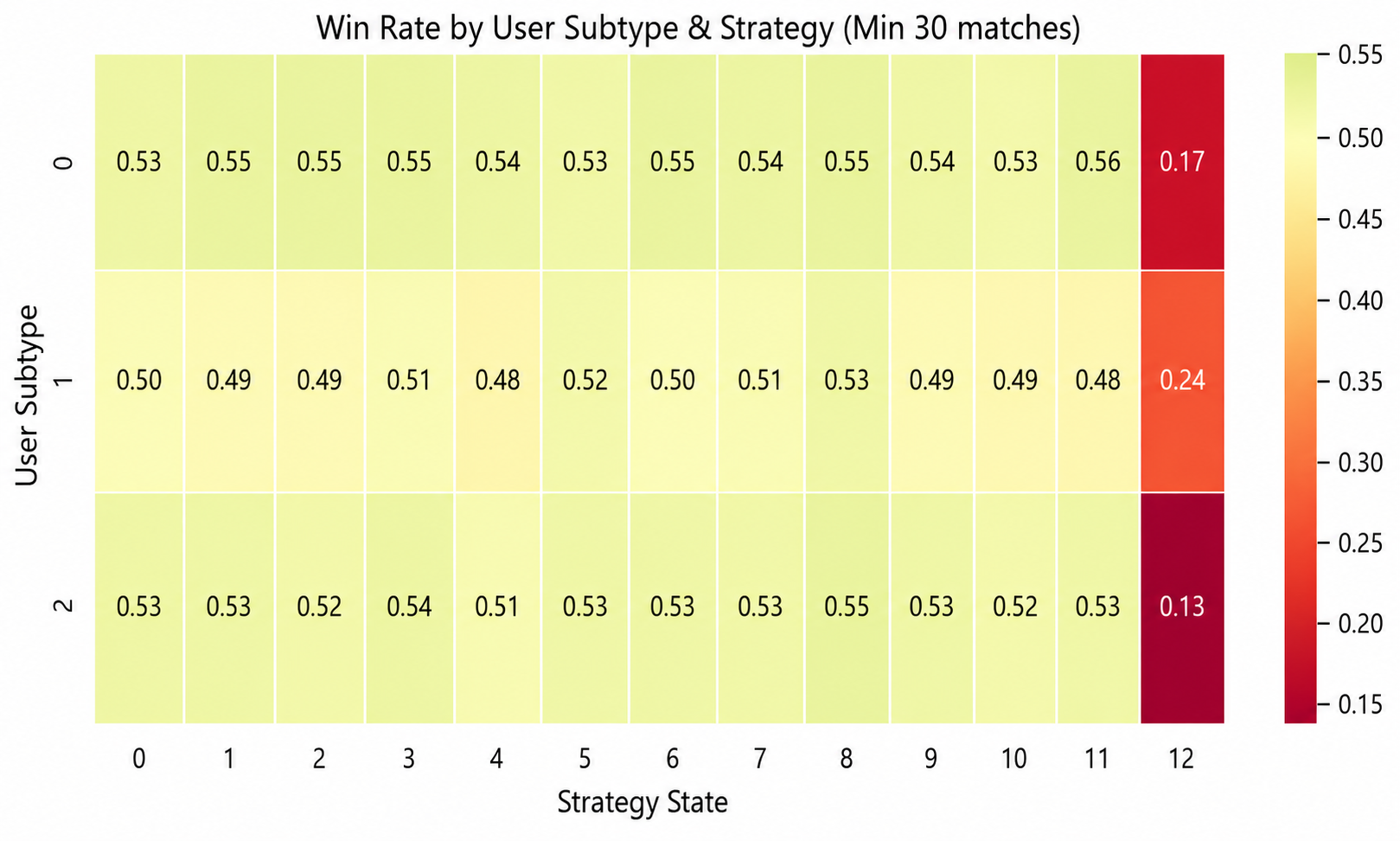}
    \caption{Win rate by user subtype and strategy state. State~12 exhibits abnormally low win rates across all subtypes, indicating a structural disadvantage under the current meta.}
    \label{fig:winrate}
\end{figure}

\textbf{Subtype~0: One-deck Loyalist (48.1\%).}
Representing nearly half of the population, this group maintains its strategy regardless of match outcomes, with an \texttt{overall\_switch\_rate} of approximately $0.00$ and near-zero transition rates after both wins and losses. Despite minimal strategic variation, they record the highest win rates of $0.53$--$0.56$ across most states (Figure~\ref{fig:winrate}). This pattern is consistent with the hypothesis that strategy repetition facilitates mastery accumulation, which in turn supports higher win rates. For Subtype~0, strategic consistency appears to function as a performance advantage: mastery accumulated through repeated use of a single strategy is associated with sustained win-rate benefits. Consequently, recommending a strategy switch to this group would lack empirical justification. It could also disrupt a behaviorally stable pattern that is associated with superior outcomes.

\begin{figure}[h]
    \centering
    \includegraphics[width=0.75\textwidth]{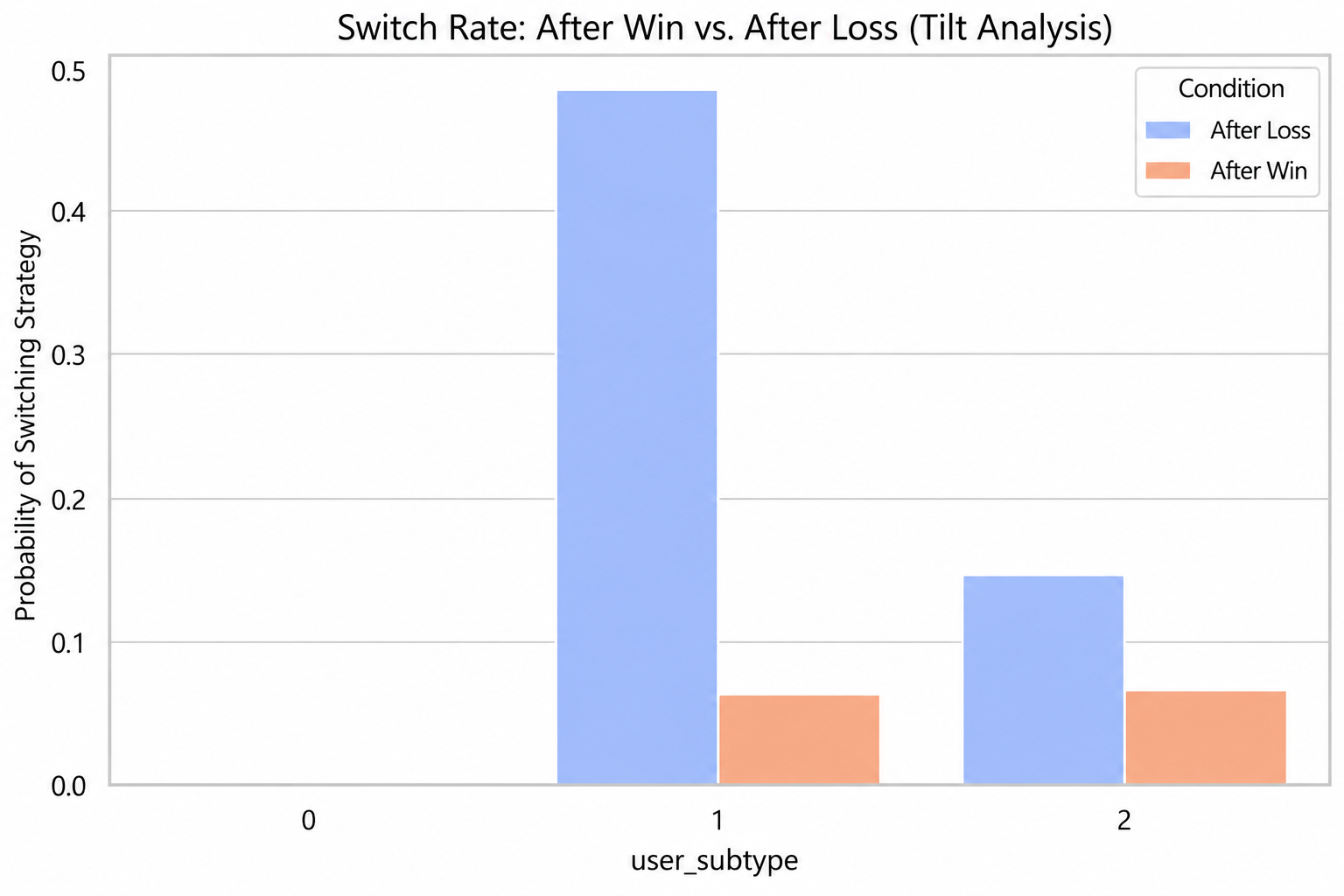}
    \caption{Switch rate after loss vs.\ after win by subtype. Subtype~1 exhibits post-loss transition rates that are $7.6\times$ higher than post-win rates ($0.487$ vs.\ $0.064$), consistent with tilt-like reactivity.}
    \label{fig:tilt}
\end{figure}

\textbf{Subtype~1: Loss-Reactive Switcher (16.0\%).}
This group exhibits the highest transition frequency, with an \texttt{overall\_switch\_rate} of 0.27. Their post-loss transition rate reaches 0.487, approximately 7.6 times higher than their post-win rate of 0.064. This asymmetric tendency to switch following losses is captured by \texttt{loss\_reactivity} = 0.423 (Figure~\ref{fig:tilt}). However, despite this high exploratory activity, their state-specific win rates are the lowest among all groups at $0.48$--$0.53$ (Figure~\ref{fig:winrate}), trailing Subtype~0 by $3$--$5$ percentage points on average. This is a substantial gap in a competitive context. Their transitions show repetitive cycles between structurally similar archetypes (e.g., State~$0 \leftrightarrow 6$), indicating broad exploration of the strategy space without corresponding performance gains.

\textbf{Subtype~2: Flex Player (35.9\%).}
This group selectively transitions while maintaining a core strategy, with an \texttt{overall\_switch\_rate} of $0.10$ and a \texttt{loss\_reactivity} of $0.082$. As shown in Figure~\ref{fig:tilt}, the difference between their post-loss transition rate of 0.148 and post-win transition rate of 0.066 is substantially smaller than that of the Loss-Reactive Switcher, indicating that their switching behavior is less reactive to match outcomes. Their win rates are stable at $0.51$--$0.55$, and their transition behavior follows a pattern we term "Basecamp," in which players maintain a primary strategy while making occasional deliberate excursions to adjacent archetypes.

\textbf{Behavioral Divergence and Pipeline Motivation.}
This three-way behavioral divergence maps directly onto the three stages of the TQP pipeline: PersonaGate (\textit{Who}), TimingGate (\textit{When}), and ScoreFusion (\textit{What}), which are described in detail in Section~\ref{sec:method}.

\section{Method}
\label{sec:method}
\subsection{Problem Formulation}
\label{sec:formulation}

The optimization goal of this study is defined as the change in win rate following a strategic transition, denoted as $\Delta WR$:

\begin{equation}
    y_{tq} = WR_{\text{next}} - WR_{\text{current}}
    \label{eq:ytq}
\end{equation}

where $WR_{\text{current}}$ represents the player's recent win rate at the moment of transition, and $WR_{\text{next}}$ denotes the observed win rate following the transition.

However, optimizing solely for $y_{tq}$ is insufficient. Players experiencing losing streaks often exhibit a natural tendency to recover their win rates over time, regardless of whether they implement a strategic change. If $y_{tq}$ is utilized as the sole optimization target, the model may indiscriminately recommend transitions to any player on a losing streak, irrespective of the transition's actual utility. This is a direct manifestation of the fundamental flaw in the Zero Switching Cost Assumption.
To approximate the net effect of switching relative to staying under matched conditions, we define the Net Transition Effect as follows:

\begin{equation}
    \Delta WR_{\text{net}} = y_{tq} - \texttt{stay\_baseline}(s, u, b)
    \label{eq:net}
\end{equation}

where $\texttt{stay\_baseline}(s, u, b)$ is the average $y_{tq}$ observed among players who did \emph{not} switch strategy, matched to the same strategy state $s$, behavioral subtype $u$, and win-rate bucket $b$. Intuitively, it answers the question: ``how well do players like this typically perform when they simply stay?''
Consequently, the decision criterion shifts from ``Did the win rate increase after the transition?'' to ``Did switching actually outperform simply staying with the current strategy?'' A transition is deemed beneficial only when $\Delta WR_{\text{net}} > 0$. This formalization serves as the primary supervisory signal for TimingGate and informs candidate ranking in ScoreFusion. Specifically, TimingGate determines whether switching is preferable to staying at the current moment using the net transition effect ($\Delta WR_{\text{net}}$), whereas TQP is trained to predict the raw post-transition gain ($\hat{y}_{tq} = WR_{\text{next}} - WR_{\text{current}}$) and is used in ScoreFusion to rank candidate destinations among moments that have already passed the timing gate.

Building on this formulation, we instantiate the transition decision problem as a three-stage pipeline:  \textsc{Who}~$\rightarrow$~\textsc{When}~$\rightarrow$~\textsc{What}. PersonaGate (Section~\ref{sec:persona}) restricts the pipeline to players for whom transitions are empirically beneficial, TimingGate (Section~\ref{sec:timing}) identifies moments when $\Delta WR_{\text{net}} > 0$, and ScoreFusion (Section~\ref{sec:fusion}) ranks candidate strategies by combining a CatBoost-based adoptability signal with TQP-predicted quality. All three stages share the player representations produced by the Player State Encoder (Section~\ref{sec:encoder}).

\subsection{Player State Encoder}
\label{sec:encoder}

Section~\ref{sec:analysis} shows that behavioral subtypes differ markedly in their responses to defeat: Loss-Reactive Switchers exhibit a transient spike in switching immediately after a loss, yet attain the lowest long-term win rates. This dissociation between short-term reactivity and stable behavioral tendency motivates a player representation that explicitly captures both aspects. To this end, we design a GRU-based encoder with multi-task pre-training, using the Strategy States and Behavioral Subtypes defined in Section~\ref{sec:analysis} as supervisory signals. We adopt a bidirectional GRU to enrich the session-level representation by jointly modeling forward temporal dynamics and backward contextual dependencies, while preserving the sequential inductive bias of recurrent architectures.

\textbf{(1) Input and Sequence Modeling.}
The encoder takes the $K=10$ most recent matches as input. Categorical variables, including Strategy State $s_t$, Win/Loss $w_t$, Deck Change $dc_t$, and Crown Difference $cd_t$, are embedded into a continuous space. Continuous variables, namely Inter-game Time Gap $\Delta t$ and Average Elixir $e_t$, are processed through Layer Normalization followed by linear projection. The deck is represented as the mean of its card embeddings. A bidirectional GRU ($\text{hidden}=128$, $\text{layers}=2$) encodes the sequence, and the final forward and backward hidden states are concatenated to produce the session representation:
\begin{equation}
    z_{raw} = W_o \, [\,\overrightarrow{h}_K \,\|\, \overleftarrow{h}_1\,]
    \label{eq:zraw}
\end{equation}

\textbf{(2) Mastery Feature Injection.}
While the GRU captures temporal patterns within the sequence, it is less suited to representing window-level summary statistics such as overall win rate or recent behavioral shifts. To provide these global summaries explicitly, we introduce a 7-dimensional Mastery Feature Vector ($mf$), consisting of average win rate, average deck switch rate, average elixir, tilt signal, win-rate trend, crown-score trend, and deck-switch concentration. The tilt signal is defined as the number of losses in the three most recent matches, yielding an integer in $\{0,1,2,3\}$. This window-based formulation captures recent loss accumulation without requiring strictly consecutive defeats. The resulting vector is incorporated through a learned linear projection:
\begin{equation}
    z_{cls} = z_{raw} + \text{MasteryProj}(mf)
    \label{eq:zcls}
\end{equation}
The long-term user representation $z_{user}$ is computed as an exponentially decayed average of prior-session $z_{cls}$ vectors (decay $= 0.9$), assigning greater weight to recent history while avoiding future data leakage.

\textbf{(3) Multi-task Pre-training.}
Five tasks are trained concurrently to guide the encoder toward representations that capture both behavioral identity and transition quality. Among them, Deck Change Prediction (\texttt{head\_dc}) and Transition Type Classification (\texttt{head\_dv}) are assigned higher loss weights. The \texttt{head\_dv} module classifies each window into one of three transition types: no deck change, within-state card adjustment, or full cross-state strategy overhaul. In doing so, it explicitly encodes the structural depth of a transition, while \texttt{head\_dc} provides the binary switching signal that anchors the temporal dynamics of strategic change. Together, these two objectives capture both the occurrence and the nature of transitions, which are the two aspects most central to TQP's recommendation objective. The remaining tasks, Win/Loss Prediction (\texttt{head\_win}), Behavioral Subtype Classification (\texttt{head\_sub}), and Crown Difference Regression (\texttt{head\_cd}), serve as auxiliary objectives that regularize the encoder toward broader performance and identity signals. After pre-training, the encoder is frozen and used as the shared backbone for TQP and TimingGate.

\subsection{PersonaGate (\textsc{Who})}
\label{sec:persona}
PersonaGate serves as the coverage-control stage of the pipeline. It restricts recommendations to players for whom transitions are supported by empirical evidence of potential benefit. Players whose behavioral consistency is associated with superior outcomes are excluded from recommendation, thereby avoiding potentially harmful interventions for this group. It uses the behavioral subtype labels pre-assigned through the clustering analysis in Section~\ref{sec:subtypes} as the direct criterion for selecting target players. One-deck Loyalists are excluded and receive an immediate \textit{Stay} decision, while Flex Players and Loss-Reactive Switchers are forwarded to the subsequent stages.

\subsection{TimingGate (\textsc{When})}
\label{sec:timing}
TimingGate is a binary classifier designed to assess whether the current moment exhibits contextual signals consistent with a beneficial transition decision. Specifically, it evaluates whether the player's recent behavioral state resembles conditions under which transitions have historically outperformed staying among matched players. By formulating timing prediction as an independent learning problem, the system enhances the precision of its recommendations.

\textbf{Performance-Oriented Labeling Strategy.}
A positive label (label$=1$) is assigned not merely when a player switches strategies, but specifically when the transition outperforms staying, as quantified by $\Delta WR_{\text{net}} > 0$ (Equation~\ref{eq:net}). This criterion ensures that a positive label is assigned only when switching genuinely outperforms staying, excluding cases in which win-rate improvement would have occurred regardless of the strategic change.

During training, Stay samples are undersampled to match the number of Switch samples. Without this, a classifier trained only on Switch samples would never observe the feature distribution of Stay situations, leading to unreliable predictions in practice, where Stay contexts constitute the majority. Negative samples are drawn from two sources: transitions that resulted in below-baseline outcomes ($\Delta WR_{\text{net}} \leq 0$), and undersampled Stay instances matched in count to the total Switch pool. This ensures that the model is exposed to the feature distribution of Stay contexts during training. Undersampling is applied only to the training set. Evaluation on the validation and test sets is conducted on the full unmodified distribution. In all splits, a transition is labeled positive only when it outperforms staying under the same conditions, ensuring label consistency throughout. Residual class imbalance within the training mix is further addressed via a positive class weight ($w_{pos}=1.5$) in the binary cross-entropy loss of TimingGate, independent of the multi-task loss weights used during encoder pre-training.

Both the quality predictor (TQP MLP) and TimingGate are implemented as 3-layer MLPs with a hidden dimension of 256.

\subsection{ScoreFusion (\textsc{What})}
\label{sec:fusion}

Once PersonaGate and TimingGate have approved a player and a moment, ScoreFusion completes the intervention decision by resolving which transition target can be recommended as an actionable intervention. To preserve the complementarity between feasibility and expected benefit, ScoreFusion separates the two objectives: the CatBoost Ranker learns relative adoptability from historical selection patterns, while TQP provides an independent estimate of the expected $\Delta WR$ for each candidate. ScoreFusion thus operationalizes the \textsc{What} decision as the selection of a transition that is both behaviorally feasible for the current player and predicted to yield a positive transition effect ($\hat{y}_{tq} > 0$).

\vspace{0.5em}
\noindent\textbf{Candidate Filtering.}
The candidate pool is constructed in two stages.
First, only transitions observed at least three times for the same behavioral subtype and current strategy state (\texttt{from\_state})
in the training data are retained, filtering out rare and noisy transitions. Second, any candidate with $TQP \leq 0$ is discarded, as it implies no expected win rate improvement. If no valid candidates remain, the system defaults to \textit{Stay}.

\vspace{0.5em}
\noindent\textbf{Score Fusion.}
Raw TQP values (mean $\approx$ 0.0004, std $= 0.106$) are orders of magnitude smaller than CatBoost outputs in $[0,\,1]$. Without rescaling, the mixing weight $\alpha$ becomes ineffective. We apply $\tanh$ with scale parameter $0.1$, chosen to match the empirical spread of TQP predictions: $\tanh(\pm 0.1\,/\,0.1) \approx \pm 0.76$, bringing both signals into a comparable range. CatBoost scores are min-max normalized to $[0,\,1]$ per query. The fused score is:
\begin{equation}
    \text{score}(s') = \alpha \cdot \text{norm\_CB}(s')
    + (1 - \alpha) \cdot \tanh\!\left(\frac{TQP(s')}{0.1}\right)
\end{equation}
The mixing weight $\alpha$ is selected on the validation set via grid search. The optimal value of $\alpha = 0.5$ assigns equal weight to both signals, confirming that adoptability and quality are complementary: neither alone achieves the performance of their combination.

\section{Results}

\subsection{Experimental Setup}

\textbf{Dataset and Splits.}
The dataset described in Section~\ref{sec:data} yields 409,293 strategy transition sequence samples after applying a sliding window of $K=10$, partitioned per player into training (80\%), validation (10\%), and test (10\%) sets. PersonaGate excludes One-deck Loyalists from recommendations; policy evaluation is therefore conducted only on Loss-Reactive Switchers and Flex Players.

\textbf{Baselines.}
We compare the full TQP pipeline against six baseline policies. \textit{Always-Stay} and \textit{Always-Switch} serve as trivial bounds. \textit{Win-Rate Threshold} recommends a randomly selected candidate strategy whenever the player's current win rate falls below $0.45$, without using any learned model. \textit{Population Oracle} recommends the strategy with the highest population-level average win rate. \textit{Collaborative Filtering} models player--strategy interactions using a win-rate-conditioned matrix. \textit{Last-K Transition} recommends each player's historically most frequent transition target for a given origin state and behavioral subtype, serving as a naive sequential baseline that captures personal preference without timing or quality signals.

\subsection{TQP Predictor Performance}
\label{sec:tqp_perf}
The shared encoder achieves an AUC of $0.843$ on deck change prediction, $69.7\%$ accuracy on behavioral subtype identification, and $81.0\%$ accuracy on transition type classification. These results reflect the complementary design of the five pre-training tasks: the transition-oriented tasks directly target the occurrence and structural depth of strategic changes, the identity task encourages the encoder to internalize behavioral differences across player types, and the performance tasks ground the representations in observable match-level outcomes. Together, these results confirm that the learned embeddings capture both strategic and behavioral signals before being passed to TQP and TimingGate.

To establish the reliability of TQP prior to policy deployment, we evaluate the model across three complementary dimensions.

\textbf{Prediction Accuracy.}
\begin{table}[h]
\centering
\small
\caption{TQP predictor performance on the test set.
``---'' denotes metrics not applicable to Always-Zero.}
\label{tab:tqp_predictor}
\begin{tabular}{@{}lccccc@{}}
\toprule
\textbf{Model} & \textbf{MAE} & \textbf{Dir. Acc.} & \textbf{Prec.} & \textbf{Rec.} & \textbf{F1} \\
\midrule
Always-Zero         & 0.0806 & 64.0\% & ---    & ---    & ---   \\
Logistic Regression & 0.0944 & 67.1\% & 59.8\% & 26.4\% & 0.366 \\
TQP-Transformer     & 0.0566 & 75.0\% & 61.0\% & 87.3\% & 0.718 \\
\midrule
\textbf{TQP-GRU (Ours)} 
& \textbf{0.0517} & \textbf{76.9\%} 
& \textbf{64.5\%} & \textbf{90.7\%} & \textbf{0.753} \\
\bottomrule
\end{tabular}
\end{table}
We evaluate $\Delta WR$ prediction using two metrics: MAE measures how closely the predicted win-rate change approximates the actual post-transition win-rate change, and Direction Accuracy measures whether the model correctly predicts the direction of that change, i.e., whether the transition will ultimately increase or decrease the player's win rate. Table~\ref{tab:tqp_predictor} compares TQP-GRU against three baselines of increasing complexity. \textit{Always-Zero} predicts no win-rate change for all transitions (MAE$=0.0806$, Dir. Acc.$=64.0\%$), serving as a lower bound. \textit{Logistic Regression} adds transition-level features without a learned encoder (MAE$=0.0944$, Dir. Acc.$=67.1\%$), performing worse than Always-Zero on MAE and confirming that raw features alone are insufficient. \textit{TQP-Transformer} shares the same pre-training and MLP structure as our model but uses a Transformer encoder (MAE$=0.0566$, Dir. Acc.$=75.0\%$), establishing a strong architecture-controlled baseline. TQP-GRU outperforms all three baselines with MAE$=0.0517$ and Dir. Acc.$=76.9\%$, demonstrating that the GRU encoder provides a consistent advantage over both non-learned features and Transformer-based encoding for sequential behavioral data.

\textbf{Discrimination Analysis.}
We verify that TQP can reliably distinguish beneficial transitions from harmful ones. A transition is beneficial if it increases the player's win rate ($\Delta WR > 0$), and harmful if it decreases it ($\Delta WR \leq 0$). Among the $2{,}554$ switching events in the test set, transitions that TQP predicted as beneficial achieved a mean actual win-rate gain of $+7.16$\%p, while those predicted as harmful resulted in a mean loss of $-5.76$\%p. The resulting gap of $+12.9$\%p means that transitions predicted by TQP as beneficial yield a win-rate improvement roughly $12.9$\%p higher than transitions predicted as harmful. A bootstrap test confirms that this gap is statistically significant (95\% CI: [$+12.3$\%p, $+13.6$\%p]; $p < 0.001$).

\textbf{Classification Performance.}
Under standard binary classification metrics, TQP achieves a Precision of $64.5\%$, Recall of $90.7\%$, and F1 of $0.753$. The high Recall indicates that TQP successfully captures the majority of genuinely beneficial transitions. Furthermore, among the 1,139 transitions predicted as harmful, $1{,}045$ ($91.7\%$) were confirmed to yield negative outcomes ($\Delta WR \leq 0$), demonstrating that TQP effectively identifies detrimental transitions.

Taken together, these three evaluations confirm that TQP provides a statistically reliable signal for transition quality. In Section~\ref{sec:policy_comparison}, we examine the prescriptive effect of applying this predictor as a recommendation policy.

\subsection{Comparative Evaluation}
\label{sec:policy_comparison}

\textbf{Evaluation Metric.}
We adopt SwitchGap as the primary evaluation metric. SwitchGap is defined as the difference in mean $\Delta WR$ between transitions the policy approves and those it rejects, computed exclusively among actual switchers in the test set:
\begin{equation}
\begin{split}
\text{SwitchGap} &=
    \mathbb{E}[\Delta WR \mid \hat{a}=\text{sw},\, a=\text{sw}] \\
  &\quad - \mathbb{E}[\Delta WR \mid \hat{a}=\text{stay},\, a=\text{sw}]
\end{split}
\label{eq:switchgap}
\end{equation}
where $\hat{a}$ denotes the policy's recommendation, and $a=\text{sw}$ restricts the population to actual switchers. Unlike Recall@K and NDCG@K, SwitchGap does not treat observed transitions as ground truth, which is critical given that frequent switchers exhibit the lowest win rates in our dataset. We additionally report \textit{Rec\_TQP}, the mean TQP-predicted $\Delta WR$ among approved transitions, which measures the expected quality of recommended transitions, and \textit{Prec@1}, the fraction of approved transitions that yield a positive actual $\Delta WR$.

\textbf{Policy and Ablation Comparison.}
\begin{table}[t]
\centering
\footnotesize
\setlength{\tabcolsep}{4pt}
\renewcommand{\arraystretch}{0.95}
\caption{Policy comparison and incremental ablation results on the test set.
``---'' indicates that the metric is undefined when all events are uniformly approved or rejected.}
\label{tab:policy}
\begin{tabular}{@{}lcccc@{}}
\toprule
\textbf{Policy / Configuration} & \textbf{Sw\%} 
  & \textbf{SwitchGap} & \textbf{Rec\_TQP} & \textbf{Prec@1} \\
\midrule
\multicolumn{5}{@{}l}{\textit{Baselines}} \\
\midrule
Always-Stay          & 0.0\%   & ---       & ---       & ---    \\
Always-Switch        & 100.0\% & ---       & $+0.9$\%p & 39.4\% \\
WR-Threshold         & 34.8\%  & $+5.2$\%p & $+4.5$\%p & 51.8\% \\
Pop.\ Oracle         & 7.6\%   & $+0.8$\%p & $+0.2$\%p & 44.1\% \\
Collab.\ Filter      & 63.6\%  & $+4.7$\%p & $+2.4$\%p & 43.8\% \\
Last-K Trans.        & 100.0\% & ---       & $+0.7$\%p & 39.4\% \\
\midrule
\multicolumn{5}{@{}l}{\textit{Incremental Pipeline (Ablation)}} \\
\midrule
(a) CatBoost only    & 100.0\% & ---         & $+1.4$\%p  & 39.4\% \\
(b) + PersonaGate    & 100.0\%$^\dagger$ 
                              & ---         & $+1.4$\%p  & 39.4\% \\
(c) + TimingGate     & 6.0\%   & $+8.24$\%p & $+8.22$\%p & 60.3\% \\
\textbf{(d) + TQP Fusion} 
                     & \textbf{5.4\%}
                     & \textbf{+10.4\%p}
                     & \textbf{+9.8\%p}
                     & \textbf{70.4\%} \\
\bottomrule
\end{tabular}

\vspace{0.3em}
\begin{minipage}{0.95\textwidth}
\footnotesize
$^\dagger$ One-deck Loyalists (48.1\%) are excluded by PersonaGate and are absent from the test set by construction.
\end{minipage}
\end{table}
Table~\ref{tab:policy} compares the baselines and incremental pipeline configurations under a unified evaluation framework. Among the baselines, high-frequency policies approve transitions indiscriminately, whereas selective policies lack the quality signal needed to identify truly beneficial ones. WR-Threshold captures a real signal but fails to assess whether the current moment is decision-ready (SwitchGap $+5.2$\%p at $34.8\%$ coverage), while Population Oracle ignores individual behavioral context entirely (+0.8\%p at $7.6\%$). Collaborative Filtering yields the highest coverage among non-trivial baselines ($63.6\%$) but achieves only +4.7\%p.

The incremental ablation reveals how each pipeline stage contributes to this improvement. Configuration~(a) approves all transitions indiscriminately and therefore offers no discriminative signal. Configuration~(b) adds PersonaGate, excluding One-deck Loyalists ($48.1\%$) from recommendations; its contribution is reflected in population-level coverage rather than SwitchGap, by construction. The critical improvement occurs at configuration~(c), where TimingGate alone achieves a SwitchGap of $+8.24$\%p at a recommendation rate of 6.0\%, demonstrating that the learned timing classifier effectively identifies decision-ready moments. Adding TQP Fusion in configuration~(d) further raises SwitchGap to $+10.4$\%p while reducing the recommendation rate to 5.4\%, confirming the complementary discriminative advantage of the quality signal. TQP's Prec@1 of $70.4\%$ represents a $+31$\%p improvement over the base rate of $39.4\%$ observed under Always-Switch.

\textbf{Ablation Study.}
We further isolate the contribution of each input representation within TQP. Removing $z_\text{cls}$ reduces direction accuracy from 76.9\% to 70.3\%, confirming that sequential match context is the primary driver of transition quality prediction. Removing $z_\text{user}$ reduces accuracy to 75.9\%, indicating that personalized behavioral history provides an orthogonal and complementary signal.

\subsection{Subtype Analysis}
Within-subtype evaluation shows that the pipeline achieves a SwitchGap of $+10.55$\%p among Loss-Reactive Switchers and $+10.23$\%p among Flex Players. In particular, the highest gain is observed in Loss-Reactive Switchers, whose post-loss transition rate of $0.487$ coexists with the lowest mean win rate among subtypes, indicating that the pipeline delivers its largest benefit to the group most exposed to unproductive loss-triggered switching. Mechanically, TimingGate withholds recommendations at the majority of moments for this group, whereas for Flex Players it approves transitions more conservatively, reflecting that the behavioral context of selective switchers already encodes a degree of strategic deliberateness. The near-identical magnitudes across two behaviorally distinct subtypes further suggest that the \textsc{Who}$\rightarrow$\textsc{When}$\rightarrow$\textsc{What} design provides discriminative quality independent of a player's switching tendency.

\section{Future Work}
This work reflects several deliberate design choices, each of which suggests promising directions for future research.

First, the proposed framework demonstrates that meaningful transition signals can be extracted from a player's behavioral history alone. Incorporating lightweight opponent-side signals, such as deck-archetype frequencies within a given trophy range, could further refine transition timing.

Second, the behavioral subtypes in this work are derived from a player's full match history, providing stable and reproducible labels that support consistent training across the pipeline. However, this static assignment cannot capture players whose behavioral tendencies shift over time. As a partial mitigation, subtype labels could be recomputed periodically over rolling windows of recent matches, allowing the pipeline to adapt to gradual behavioral drift while preserving the stability of offline label assignment. A fully online model that tracks such changes in real time would further extend this direction for players whose play style is still evolving.

Third, while reinforcement learning research identifies optimal strategies for an agent, it typically does not model whether a human player is ready to adopt those strategies. TQP complements this by modeling switching costs and individual behavioral tendencies. It could therefore serve as a practical bridge between RL-generated strategy recommendations and real player behavior.

Fourth, SwitchGap is grounded in observed post-transition outcomes, providing a direct measure of whether a policy distinguishes between beneficial and detrimental transitions without relying on assumed ground truth. This design enables meaningful policy comparison even in settings where observed player choices are systematically suboptimal. Online evaluation, in which the pipeline directly guides player decisions, would further validate the system's prescriptive value beyond what offline metrics can provide.

Fifth, several broader directions remain open, including validating the framework across other competitive games and genres, adapting recommendations to players at different skill levels, and improving the interpretability of recommended transitions.

\section{Conclusion}
This paper investigates strategy-switching behavior in competitive mobile gaming through an analysis of 926,334 match records from 34,619 \textit{Clash Royale} players. The analysis reveals a central pattern: players who switch strategies most frequently (Loss-Reactive Switcher: \texttt{overall\_switch\_rate} $= 0.27$) achieve the lowest win rates ($0.48$--$0.53$), whereas well-timed transitions are associated with measurable positive effects. This finding suggests that performance is determined not by switching frequency itself, but by the timing and conditions under which transitions occur, a problem that existing recommendation frameworks have largely overlooked due to their implicit assumption of zero switching cost.

We address this issue by instantiating the transition decision problem as a three-stage pipeline comprising PersonaGate, TimingGate, and ScoreFusion, with TQP providing the transition-quality signal in the WHAT stage. The pipeline achieves a SwitchGap of +10.4\%p at only 5.4\% coverage and a Prec@1 of 70.4\%, substantially outperforming all baselines. Ablation studies confirm the incremental contribution of each stage.

The finding that frequent switchers achieve the worst outcomes is not merely descriptive; it is also prescriptive. The players who switch most frequently after losses are precisely those who stand to benefit the most from timely and well-resolved guidance. Effective transition recommendation therefore requires more than identifying what to play next. It requires a decision-centric formulation under behavioral constraints. Concretely, this involves determining who should be guided, when intervention is warranted, and what transition can be recommended as a feasible, player-conditioned, and performance-improving action.

\bibliographystyle{unsrt}  
\bibliography{references}

\end{document}